\documentclass[10pt,twocolumn,letterpaper]{article}

\usepackage{cvpr}              
\usepackage{array}
\usepackage{multirow}
\usepackage{dsfont}
\usepackage{bbding}
\usepackage[normalem]{ulem}
\usepackage{balance}
\usepackage{wasysym}
\usepackage{amssymb}
\usepackage{pifont}
 
\definecolor{cvprblue}{rgb}{0.21,0.49,0.74}
\usepackage[pagebackref,breaklinks,colorlinks,allcolors=cvprblue]{hyperref}

\def\confName{CVPR}
\def\confYear{2026}

\title{Training-free Motion Factorization for Compositional Video Generation}

\author{
Zixuan Wang$^{1}$ \quad
Ziqin Zhou$^{2}$ \quad
Feng Chen$^{2}$ \quad
Duo Peng$^{3}$ \quad
Yixin Hu$^{1}$ \quad \\
Changsheng Li$^{4}$ \quad
Yinjie Lei$^{1, *} $
\\
$^1$Sichuan University \quad
$^2$The University of Adelaide \\
$^3$Nanyang Technological University \quad
$^4$Beijing Institute of Technology \\
{\tt\small zixuan98@stu.scu.edu.cn, yinjie@scu.edu.cn} \\
}
\begin{document}
\twocolumn[{%
\renewcommand\twocolumn[1][]{#1}%
\maketitle

\begin{center}
 \includegraphics[width=1.00\linewidth]{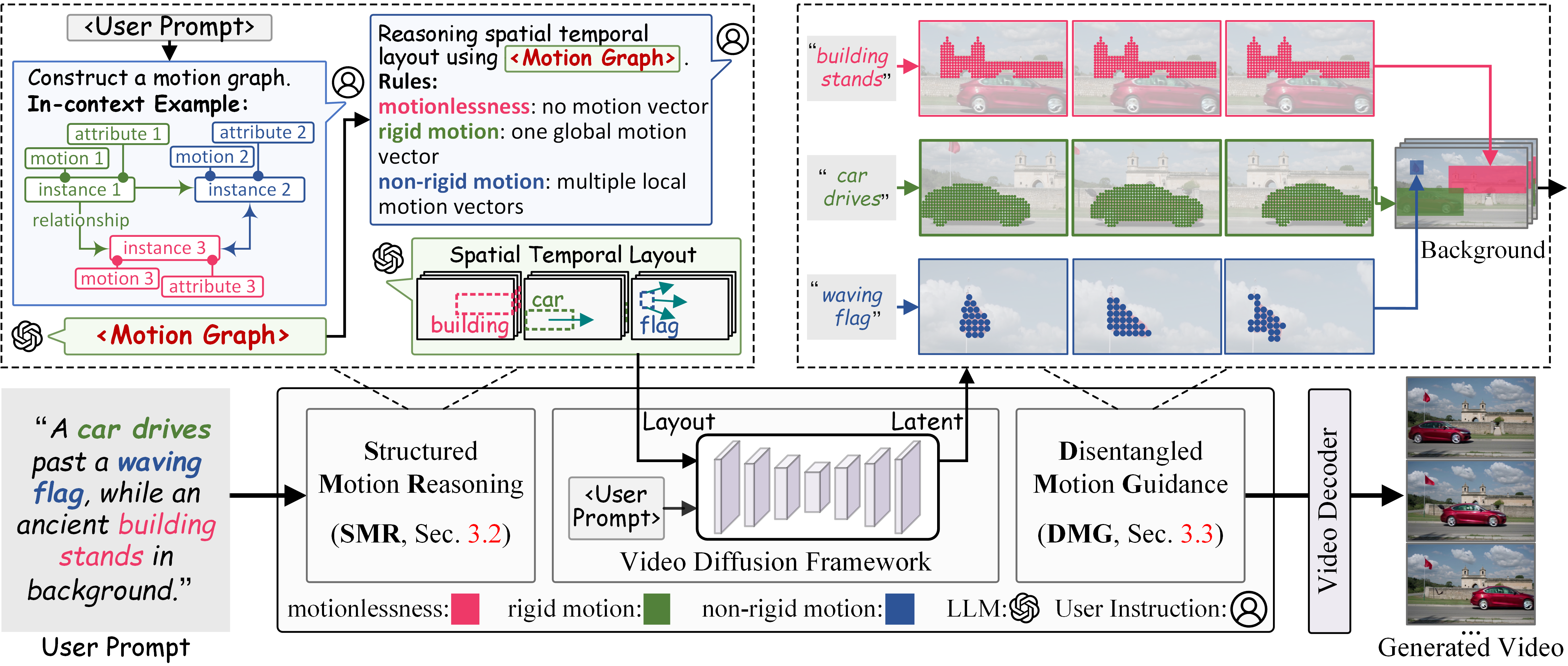}
 \captionof{figure}{Overview of our motion factorization framework. First, for each instance belonging to a particular motion category, our framework infers its per-frame changes in shape and position from a structured motion graph (Sec.~\ref{cmpa}). Second, conditioned on the motion category, dedicated guidance branches synthesize per-instance motions, which are subsequently composed into a coherent scene (Sec.~\ref{cdmg}).
}
 \label{fig:main_figure}
\end{center}
}]

\insert\footins{\noindent\footnotesize\textsuperscript{*} Corresponding author.}
\begin{abstract}
Compositional video generation aims to synthesize multiple instances with diverse appearance and motion. However, current approaches mainly focus on binding semantics, neglecting to understand diverse motion categories specified in prompts. In this paper, we propose a \textbf{motion factorization} framework that decomposes complex motion into three primary categories: motionlessness, rigid motion, and non-rigid motion. Specifically, our framework follows a \textit{planning before generation} paradigm. (1) During planning, we reason about  motion laws on the motion graph to obtain frame-wise changes in the shape and position of each instance. This alleviates semantic ambiguities in the user prompt by organizing it into a structured representation of instances and their interactions. (2) During generation, we modulate the synthesis of distinct motion categories in a disentangled manner. Conditioned on the motion cues, guidance branches stabilize appearance in motionless regions, preserve rigid-body geometry, and regularize local non-rigid deformations. Crucially, our two modules are model-agnostic, which can be seamlessly incorporated into various diffusion model architectures. Extensive experiments demonstrate that our framework achieves impressive performance in motion synthesis on real-world benchmarks. Code is available at \textcolor[RGB]{237,0, 140}{https://github.com/ZixuanWang0525/MF-CVG}.
\end{abstract}    
\section{Introduction}
Compositional Video Generation (CVG) focuses on generating high-quality videos from complex user prompts \cite{li2024survey,lei2024comprehensive,xue2025human}. These prompts describe complex scenes with multiple interacting instances, each characterized by its own appearance and motion categories. As commercial video generation models \cite{pika2024,Runway2024,dreamina2024,kling2024} are increasingly deployed, CVG has been widely applied in real-world scenarios, such as virtual reality \cite{jin2021subjective} and human-computer interaction \cite{zuo2023fine}. However, a crucial challenge of such generation models is their inability to the diversity of motion categories. As a result, generated motions appear overly similar, even with markedly distinct user prompts.

In recent years, CVG frameworks have been widely investigated for modeling realistic motion of individual instances \cite{lian2023llm,lin2023videodirectorgpt,huang2024genmac,tian2024videotetris,zhang2025magiccomp}. These frameworks typically comprise a dual-stage process. First, a large language model (LLM) serves as the video planner to generate the sequence of bounding boxes for each instance. Second, the video generator constrains the motion of instances to follow the trajectories defined by these box sequences. However, generating diverse categories of motion remains a challenge, because (1) \textit{Motion semantics are ambiguous}. Directly generating box sequences from user prompts may cause broken motion paths and abnormal size variations, owing to linguistic ambiguity. (2) \textit{Motion guidance is rough}. Uniform diffusion guidance fails to differentiate among diverse motion categories, leading to conflated and implausible dynamics.

In this paper, we propose a motion factorization framework to improve motion diversity, as shown in ~\cref{fig:main_figure}. Specifically, we decompose motion into three primary categories (motionlessness, rigid, and non-rigid motions), with each instance uniquely assigned to a single category. Based on this categorization, we propose two modules to sequentially plan and guide motion generation. First, to resolve motion ambiguity in user prompts, we develop a Structured Motion Reasoning (SMR) module (Sec.~\ref{cmpa}). Instead of directly inferring motion representations from the prompt, our module structures the prompt into a motion graph representing instances and their interactions. This graph enables reasoning about motion laws to generate diverse spatial-temporal layouts. Second, we design a Disentangled Motion Guidance (DMG) module (Sec.\ref{cdmg}) that synthesizes diverse motion through specialized guidance branches. Static instances are anchored to a reference frame to maintain consistent appearance and position. For rigidly moving instances, we enforce geometric invariance leveraging a frame-agnostic shape template. Under non-rigid motion, flexible shape variation of each instance is modeled with a dense pixelwise displacement field.

To evaluate the performance of our framework on CVG, we construct datasets that cover more complex and diverse scenarios. The prompts in our dataset are derived from descriptions of real-world videos rather than handcrafted templates. We implement our framework on both VideoCrafter-v2.0 \cite{chen2024videocrafter2} (3D U-Net architecture) and CogVideoX-2B \cite{yang2024cogvideox} (DiT architecture), and validate its effectiveness on these datasets. Comprehensive experiments demonstrate the superiority of our approach on CVG, particularly in generating desired motion categories for each instance.

Our contributions can be summarized as follows:
\begin{itemize}
\item We enhance motion diversity in CVG by factorizing scene dynamics into three canonical categories, including motionlessness, rigid, and non-rigid motions.
\item To yield unambiguous motion representations, we introduce a structured motion graph as a bridge to reason about motion laws across diverse motion categories.
    \item To enable disentangled synthesis, we design specialized guidance branches for appearance consistency, geometric invariance, and local deformation.
\item Through extensive experiments, we show that our framework achieves superior motion generation performance across both 3D U-Net and DiT architectures.
\end{itemize}

\section{Related Works}
\textbf{Compositional Visual Generation.} To model multiple instances and relationships in complex scenes, compositional visual generation has been explored. For image, some approaches \cite{chen2024pixart,li2024hunyuan,esser2024scaling} parse compositional semantic units from complex prompts. Others \cite{xie2023boxdiff,xiao2023r,phung2024grounded} progressively update noisy embeddings during the sampling process to align with compositional bounding boxes. For video, semantics and location across time-axis should be also considered. In semantic understanding, VideoTetris \cite{tian2024videotetris} decomposes prompts at both frame and instance levels. Vico \cite{yang2024compositional} rebalances token importance of action-related words. In sampling guidance, LVD \cite{lian2023llm} and VideoDirectorGPT \cite{lin2023videodirectorgpt} use sequences of bounding boxes to capture instance displacements. However, such approaches fail to consider the diversity of motion categories, often generating overly similar motion across diverse instances.

\noindent \textbf{LLMs-Assisted Video Generation.} LLMs have made a significant impact in natural language processing due to their ability to understand open-world knowledge \cite{zhao2023survey,minaee2024large}. Some recent studies \cite{hong2023direct2v,huang2023free, oh2024mevg,lu2023flowzero,he2025dyst,lian2023llm,lin2023videodirectorgpt,zhuang2024vlogger} have utilized LLMs to assist video generation by parsing user prompts as additional guidance. For example, DirecT2V \cite{hong2023direct2v} and FreeBloom \cite{huang2023free} divide user prompts into separate prompts for each frame, enabling the generation of time-varying scenes. FlowZero \cite{lu2023flowzero} and DyST-XL \cite{he2025dyst} synthesize bounding boxes to guide the dynamic interactions between multiple objects. Vlogger \cite{zhuang2024vlogger} elaborates a user story into a script to achieve long video generation. However, the above frameworks still face challenges in reasoning about the diverse interactions among multiple instances. This is mainly because they lack structured modeling to handle the semantic ambiguity of prompts. 

\noindent \textbf{Motion Guidance in Video Generation.}
Given the crucial role of motion in video generation, some studies have endeavored to synthesize realistic dynamics by leveraging diverse motion signals. Pioneering works such as VMC \cite{jeong2024vmc}, LAMP \cite{wu2024lamp}, DrugNUWA \cite{yin2023dragnuwa} and OnlyFlow \cite{koroglu2025onlyflow} focus on replicating motion categories in reference videos. But, such imitation fails to generalize to unseen motion types. In recent years, some researchers have explored motion guidance based on user-provided sparse or dense motion fields \cite{ma2023trailblazer,ma2023trailblazer,qiu2024freetraj,burgert2025go,zhang2025tora,liang2024movideo}. TrailBlazer \cite{ma2023trailblazer} and FreeTraj \cite{qiu2024freetraj} regard bounding boxes annotated in keyframes as a sparse motion field, providing rough movement direction of a few pixels. The motion\_prompting \cite{geng2025motion} expands user-provided mouse drags into more complex semi-dense motion flows. However, these approaches generally adopt the uniform motion guidance paradigm, failing to account for the motion diversity of individual instances within a scene.

\section{Methodology}
\subsection{Overall Framework}
Our framework assigns heterogeneous motion prototypes (ranging from motionlessness to rigid and non-rigid motions) to distinct instances, conditioned on their motion categories. Specifically, in ~\cref{cmpa}, we propose a Structured Motion Reasoning (SMR) module, which represents motion categories as position and scale variations across bounding box sequences, inferred from the structured motion graph. These bounding box sequences collectively form the spatial-temporal layout.
\begin{equation}
\{ \mathcal{B}_1, \mathcal{B}_2, \dots, \mathcal{B}_F \} = \mathrm{LLM}(\mathcal{R}; C)
\end{equation}
where $\{ \mathcal{B}_1, \mathcal{B}_2, \dots, \mathcal{B}_F \}$ denotes the spatial-temporal layout. $\mathcal{R}$ specifies the structured motion graph, $C$ is the user-provided prompts. $F$ is the frame number.

Conditioned on these layouts, in ~\cref{cdmg}, we present a Disentangled Motion Guidance (DMG) module, which enables diverse motion synthesis through separate constraints on appearance consistency, geometric invariance, and spatial deformation. These branches transform spatial-temporal layout into the motion-specific masks to optimize attention maps. This process consequently updates the video embeddings $\mathbf{z}_{1:F}^{t}$. In practice, for 3D U-net architecture, we gradually update $\mathbf{z}_{1:F}^{t}$ by:
\begin{equation}
\mathbf{z}_{1:F}^{t-1} \gets \mathbf{z}_{1:F}^{t} - \nabla \mathcal{L},
\end{equation}
\begin{equation}
\mathcal{L} = 1 - \frac{\beta}{P}\sum(\mathbf{A} \odot (\mathcal{G}_{\rm m}+\mathcal{G}_{\rm r}+\mathcal{G}_{\rm nr})),
\end{equation}
where $\textbf{A}$ denotes the attention map. $P$ is the number of pixels. $\beta$ denotes the guidance factor. $\mathcal{G}_{\rm m}$, $\mathcal{G}_{\rm r}$, and $\mathcal{G}_{\rm nr}$ are masks designed to guide the generation of motionless, rigidly moving, and non-rigidly moving instances. For the DiT architecture, we directly modify the original scores:
{
\begin{equation}
\mathbf{A} = \mathrm{Softmax} \left( \frac{\mathbf{Q} \mathbf{K}^\top(1 + \beta \odot (\mathcal{G}_{\rm m}+\mathcal{G}_{\rm r}+\mathcal{G}_{\rm nr}))}{\sqrt{d}} \right).
\end{equation}
}

\begin{figure*}[t!]
    \centering
    \includegraphics[width=1.00\linewidth]{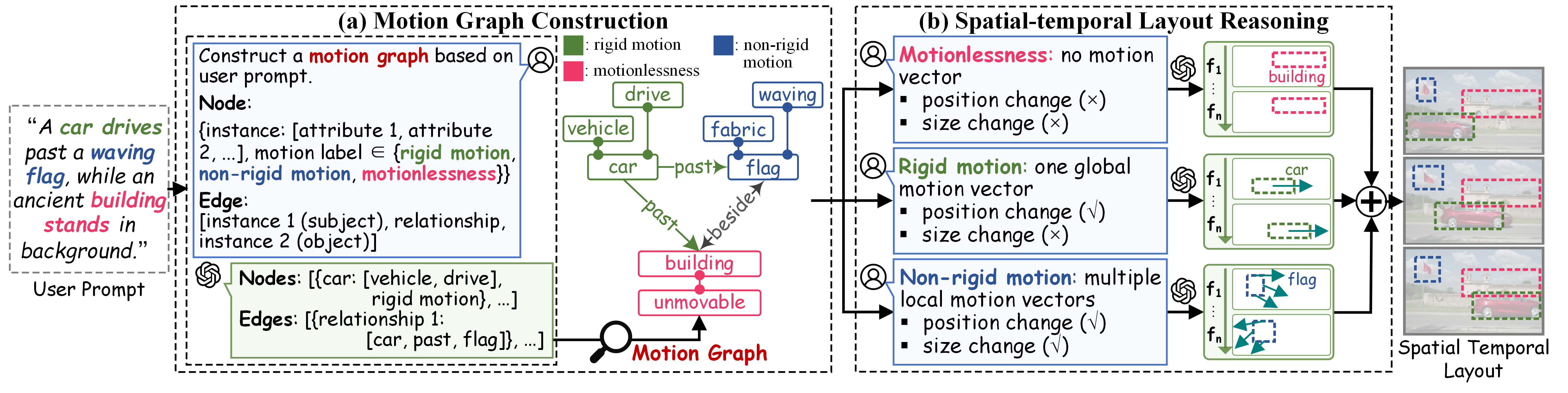}
    \caption{Overview of our Structured Motion Reasoning (SMR) module (~\cref{cmpa}). (a) Given a user prompt, we organize it into a motion graph describing instances and their interactions. (b) For each instance, conditioned on its motion category, we infer a bounding box sequence from graph-derived motion cues. All bounding box sequences are then composed into a coherent spatial-temporal layout.}
    \label{fig:cmpa}
\end{figure*}

\subsection{Structured Motion Reasoning}
\label{cmpa}
This module aims to infer motion representations that capture both individual behaviors and pairwise interactions by leveraging the semantic reasoning capability of LLMs, as shown in ~\cref{fig:cmpa}. However, user-provided prompts are often semantically ambiguous, which renders direct motion inference from such prompts unreliable. To address semantic ambiguity, we convert the original prompts into a structured motion graph, organizing instances with their associated actions and pairwise relationships. Conditioned on each instance’s motion label, we infer its bounding box sequences from motion cues derived from the motion graph. These box sequences form motion representations, encoding diverse motion categories through position and scale variations across frames.

\noindent \textbf{Motion Graph Construction.} To capture motion semantics of individual instances, we construct the motion graph through parsing compositional prompts. Specifically, each described instance is represented as a node in the graph, annotated with its corresponding motion attributes and a canonical motion label. The motion attributes are parsed by identifying verbs or predicate phrases linked to each instance. When such predicates are absent, attributes are inferred from the context. Determined by analyzing the motion attributes and instance category, each node is assigned a canonical motion label. Directed edges encode pairwise relationships between instances, including spatial relationships (\textit{e.g.}, “next to”, “on top of”) and dynamic interactions (e.g., “pass by”, “move toward”). Formally, we denote this graph as $\mathcal{R} = (\mathcal{V}, \mathcal{E})$, where $\mathcal{V}$ and $\mathcal{E}$ respectively represent the nodes and directed edges described above. 

\noindent \textbf{Spatial-temporal Layout Reasoning.} We generate motion representations for diverse categories by leveraging semantic cues encoded in constructed motion graph. Let $v_n \in \mathcal{V}$ denote the unique identifier of the $n$-th instance, and $\mathcal{B}_f(v_n)$ signify the bounding box in the $f$-th frame. For \textit{motionless instances}, the position and size of the bounding box remain unchanged across all frames; that is, $\mathcal{B}_f(v_n) = \mathcal{B}_1(v_n)$ for all $f$. For \textit{rigidly moving instance}, the position of the bounding box is updated at the current frame based on the estimated velocity $\vec{u}_{v_n}$ and acceleration $\vec{a}_{v_n}$. 
\begin{equation}
\mathcal{B}_f(v_n) = \mathcal{B}_{f-1}(v_n) + \vec{u}_{v_n} + \frac{1}{2} \vec{a}_{v_n},
\end{equation}
where $\vec{v}_{v_n}$ and $\vec{a}_{v_n}$ are predicted from the previous movement of the bounding box within a sliding window, guided by action cues in the motion graph. Unlike rigid motion that applies a single displacement vector to the entire instance, \textit{non-rigid motion} is modeled by multiple directional cues affecting distinct regions. In view of this, we infer boundary-wise displacement vectors $\Delta_f (v_n)$ by using the localized deformation implicitly captured in the motion graph. These vectors are used to update the bounding box by considering asymmetric shifts along its boundary:
\begin{equation}
\mathcal{B}_f(v_n) = \mathcal{B}_{f-1}(v_n) + \Delta_f (v_n).
\end{equation}

\begin{figure*}[tb!]
    \centering
    \includegraphics[width=1.00\linewidth]{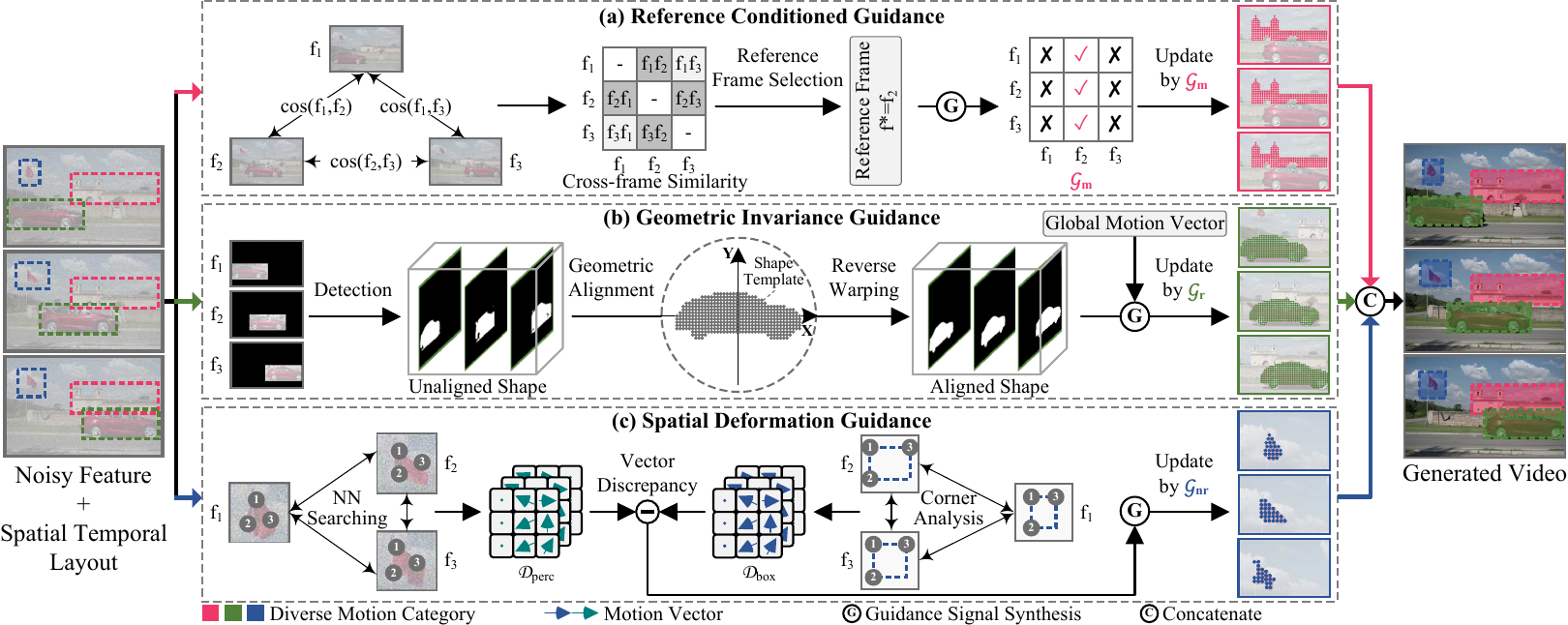}
    \caption{Overview of Disentangled Motion Guidance (DMG) module (~\cref{cdmg}). (a) For motionless instances, we enforce each frame interacts only with a designated anchor frame. (b) For rigidly moving instances, we restrict cross-frame interactions of a foreground within the shape aligned regions. (c) For instances undergoing non-rigid movements, we minimize pixel-wise discrepancies between perceptual deformations and box-induced deformations.}
    \label{fig:adaptive_motion_guidance_agent}
\end{figure*}

\subsection{Disentangled Motion Guidance}
\label{cdmg}
This module aims to enhance motion diversity during the synthesis process of video diffusion model by separately modulating motion categories of each instance. As shown in ~\cref{fig:adaptive_motion_guidance_agent}, unlike previous approaches which adopt uniform guidance across diverse motion types \cite{ma2023trailblazer, qiu2024freetraj}, we design \textit{reference conditioned guidance} to enhance cross-frame appearance consistency of motionless instances; \textit{geometric invariance guidance} to preserve geometric invariance of rigidly moving instances; \textit{spatial deformation guidance} to capture complex deformations of instances undergoing non-rigid movements.

\noindent \textbf{Reference Conditioned Guidance.} Video diffusion models often induce spurious variations in static regions, causing undesired flicker between frames. To preserve cross-frame appearance consistency, we anchor pixel-wise features in motionless regions to a stable reference frame. Specifically, we identify a reference frame $f^*$ using the minimum inter-frame feature difference criterion, under the assumption that a static instance has minimal appearance variation over time. 
\begin{equation}
f^* = \arg\min_{f} \sum_{f'=1}^{F} D\left( \varphi(\mathbf{z}_{f}^{t}), \varphi(\mathbf{z}_{f'}^{t}) \right),
\end{equation}
where $D(\cdot, \cdot)$ denotes the feature distance metric. Afterward, we achieve pixel-to-pixel appearance consistency by aligning features of all the frames to the reference frame through a masking strategy. The mask is defined as: 
\begin{equation}
\mathcal{G}_{\rm m}[x, y, f, f'](v_n) = \mathds{1} \left( f' = f^* \ \& \ (x, y) \in \mathcal{B}(v_{n}) \right),
\end{equation}
where $\mathds{1}(\cdot)$ denotes the indicator function, which returns $1$ if the condition holds and $0$ otherwise.

\noindent \textbf{Geometric Invariance Guidance.} When geometric constraints are absent, video generative models often distort rigid instances, yielding misaligned shapes. To preserve geometric invariance of rigid bodies, we confine cross-frame interactions to geometrically aligned regions derived from a frame-agnostic shape template. Specifically, we apply $k$-means clustering to separate background regions from bounding boxes, yielding foreground masks with clear shape contours. Unfortunately, such masks contain segmentation noise, leading to misaligned shapes. Given this, we aggregate coarse masks through a pixelwise consensus to generate a shape template. This template is inversely warped to each frame, producing geometrically aligned foreground masks.
\begin{equation}
\mathcal{M}_f(v_{n})
= \operatorname{Warp}\!\left(
\operatorname{Thr}\!\left\{
\operatorname{k-means}\!\left(\mathcal{B}_{f'}(v_{n})\right)
\right\}_{f'=1}^{F}
\
\right),
\end{equation}
where $\operatorname{Thr}(\cdot)$ denotes the operation of pixelwise voting. In addition, we modulate the intensity of feature interactions using the displacement magnitude between frames. This is measured by the Euclidean distance between the bounding box centers of any two frames: 
\begin{equation}
\mathbf{\Gamma}[f, f'] = \exp\left( -\alpha \cdot \|\mathbf{C}_f - \mathbf{C}_{f'}\|_2 \right) + 1,
\end{equation}
where $\mathbf{\Gamma}[f, f']$ represents the displacement penalty factor between frame $f$ and $f'$. $\alpha > 0$ is a scaling factor. $\mathbf{C}_f$ and $\mathbf{C}_{f'}$ are center coordinates of the bounding box $\mathcal{B}_{f}(v_{n})$ and $\mathcal{B}_{f'}(v_{n})$, respectively. Given both $\mathbf{\Gamma}$ and $\mathcal{M}_(v_{n})$, mask values are computed as follows: 
\begin{equation}
\mathcal{G}_{\rm r}(v_n)=\mathcal{M}(v_n) \cdot {\mathcal{M}(v_n)}^\top \odot \mathbf{\Gamma}.
\end{equation}

\noindent \textbf{Spatial Deformation Guidance.} To regularize deformation of instances undergoing non-rigid motion, we enable pixels to follow diverse motion magnitudes and directions. For such a purpose, sparse motion vectors inferred via CoT reasoning are propagated to the corresponding foreground locations, forming a fine-grained motion field. By inferred dense field, pixel-wise correspondences across frames are regularized during video generation. Specifically, we use pixel-wise nearest neighbor search across frames to obtain perceptual deformations $\mathcal{D}_{\rm perc}$, Our key insight is that cross-frame deformation can be regarded as a many-to-many pixel correspondence problem. And pixel correspondences in RGB space are approximately preserved in diffusion features \cite{geyer2023tokenflow}.
\begin{equation}
\mathcal{N}(i,j) = \arg\min_{(i', j')} 
\left\| 
\varphi\left( \mathbf{z}_{f}^{t} \right)_{i,j} - 
\varphi\left( \mathbf{z}_{f'}^{t} \right)_{i', j'} 
\right\|_2,
\end{equation}
\begin{equation}
\mathcal{D}_{\rm perc}[i,j] = \mathcal{N}(i,j) - (i,j),
\end{equation}
where $\mathcal{N}(i,j)$ denotes the matched position of pixel $(i,j)$. Thereafter, we obtain box-induced deformations $\mathcal{D}_{\rm box}$ by computing the displacement of box corners as basic motion vectors and propagating them to all positions inside the box through bilinear interpolation. 
\begin{equation}
\mathcal{D}_{\rm box}[i, j]
= \operatorname{Interp}\!\big(\{\mathbf{d}_k\}_{k=1}^{4}, (i, j)\big),
\end{equation}
where $\mathbf{d}_k$ denotes the displacement of the $k$-th box corner, and $\operatorname{Interp}(\cdot)$ specifies bilinear interpolation operation. Based on the differences between $\mathcal{D}_{\rm perc}$ and $\mathcal{D}_{\rm box}$, we modulate cross-frame correlation to follow desired non-rigid deformation as follows:
\begin{equation}
\mathbf{\Lambda}[i, j] = \exp\left(-\alpha \cdot \left(\mathcal{D}_{\rm perc}[i,j]  - \mathcal{D}_{\rm box}[i, j] \right) \right) + 1,
\end{equation}
\begin{equation}
\mathcal{G}_{{\rm nr}} = (\mathcal{M}(v_n) \cdot {\mathcal{M}(v_n)}^\top) \odot \mathbf{\Lambda},
\end{equation}
where $\mathcal{M}(v_n) $ denotes the foreground masks obtain by $k$-means clustering. $\mathbf{\Lambda}[i, j]$ represents the deformation penalty factor of the position $(i, j)$.

\newcolumntype{C}[1]{>{\centering\arraybackslash}p{#1}}
\begin{table*}[htbp!]
\centering
\caption{Performance comparison of cross-modal compositional video generation approaches on our CVGBench-m and CVGBench-p datasets. Best/2nd best scores are \textbf{bolded}/\underline{underlined}. $^\dagger$ indicates compositional generation models. }
\renewcommand{\arraystretch}{1.15}
\setlength{\heavyrulewidth}{1pt}
\resizebox{\linewidth}{!}{%
\begin{tabular}{ccccccccccc}
\toprule
\multirow{3}{*}{Models} & \multicolumn{5}{c}{CVGBench-m} & \multicolumn{5}{c}{CVGBench-p}\\ 
\cmidrule(lr){2-6} \cmidrule(lr){7-11} 
& Subject & Background & Temporal & Motion & Dynamic  & Subject & Background & Temporal & Motion & Dynamic  \\
& Consistency & Consistency & Flickering & Smoothness & Degree & Consistency & Consistency  & Flickering & Smoothness & Degree \\
\midrule
LVDM \cite{he2022latent} & 88.74\% & 91.27\%  &  89.37\%  &  91.76\% &  84.55\%  &  91.05\%  &  92.60\%  &  91.26\% &  93.79\% &  68.78\% \\
 modelScope \cite{wang2023modelscope} &  93.17\% &  93.93\% &  94.69\% &  96.23\% &  51.82\% &  95.71\% &  95.28\% &  96.01\% &  97.27\% &  30.81\% \\
 ZeroScope \cite{ZeroScope2023} &  96.41\% &  95.69\% &  97.40\% &  98.66\% &  30.77\% &  97.40\% &  96.24\% &  97.73\% &  98.84\% &  16.10\%\\
 LATTE \cite{ma2024latte} &  90.91\% &  94.33\% &  92.79\% &  94.78\% &  77.44\% &  95.21\% &  96.11\% &  95.42\% &  97.02\% &  51.66\%  \\
 VideoCrafter-v1.0 \cite{chen2023videocrafter1} &  97.06\% &  96.59\% &  96.04\% &  97.27\% &  38.16\% &  84.97\% &  92.51\% &  80.99\% &  83.02\% &  80.26\% \\
 Show-1 \cite{zhang2024show} &  95.39\% &  95.44\% &  97.37\% &  98.23\% &  31.85\% &  97.50\% &  96.47\% &  98.41\% &  98.93\% &  11.45\% \\
 LaVie \cite{wang2024lavie} &  93.22\% &  94.36\% &  93.73\% &  96.17\% &  78.90\% &  95.45\% &  95.67\% &  95.73\% &  97.38\% &  57.40\% \\
\midrule
VideoCrafter-v2.0 \cite{chen2024videocrafter2} & 97.68\% & 97.28\% & 96.28\% & 98.16\% & 33.11\% & \underline{98.30\%} & \underline{97.62\%} & 97.00\% & 98.48\% & 18.22\% \\
+ BoxDiff$^\dagger$ \cite{xie2023boxdiff} & 97.42\% &  96.93\% & 96.33\% & 98.25\% & 38.31\% & 98.08\% & 97.32\% & 96.87\% & 98.50\% & 25.44\%  \\
+ R\&B$^\dagger$ \cite{xiao2023r} & 97.37\% & 96.87\% & 96.47\% & 98.21\% & 38.35\% & 98.11\% & 97.30\% & 96.83\% & 98.56\% & 25.45\%   \\
+ A\&R$^\dagger$ \cite{phung2024grounded} & 97.48\% & 97.05\% & 96.43\% & 98.27\% & 38.40\% & 97.90\% & 97.10\% & 96.83\% & 98.47\% & 31.44\%  \\
+ Vico$^\dagger$ \cite{yang2024compositional} & \underline{97.72\%} & \underline{97.43\%} & \underline{96.68\%} & \underline{98.35\%} & \underline{40.00\%} & 98.23\% & 97.36\% & \underline{97.10\%} & \underline{98.56\%} & \underline{32.85\%}  \\
+ Ours & \textbf{98.40\%} & \textbf{98.11\%} & \textbf{97.39\%} & \textbf{98.63\%} & \textbf{82.21\%} & \textbf{98.81\%} & \textbf{98.29\%} & \textbf{97.82\%} & \textbf{98.79\%} & \textbf{78.24\%} \\
\midrule
CogVideoX-2B \cite{yang2024cogvideox} &  \underline{91.33\%} & \underline{92.78\%} & 95.01\% & 96.88\% & 87.80\% & 92.85\% & 93.32\% & 96.11\% & 97.95\% & 79.85\% \\
+ R\&P$^\dagger$ \cite{chen2024training} & 91.00\% & 90.85\% & \underline{95.07\%} & \underline{96.96\%} & \underline{91.02\%} & \underline{93.01\%} & \underline{94.26\%} & \underline{96.12\%} & \underline{97.95\%} & \underline{81.52\%}  \\
+ Ours & \textbf{98.27\%} & \textbf{97.73\%} & \textbf{98.25\%} & \textbf{98.74\%} & \textbf{96.00\%} & \textbf{98.74\%} & \textbf{98.23\%} & \textbf{98.38\%} & \textbf{98.94\%} & \textbf{87.06\%} \\
\bottomrule
\end{tabular}
}
\label{tab:video_comparison}
\end{table*}

\section{Experiments}

\begin{figure*}[tb!]
    \centering
    \includegraphics[width=1.00\linewidth]{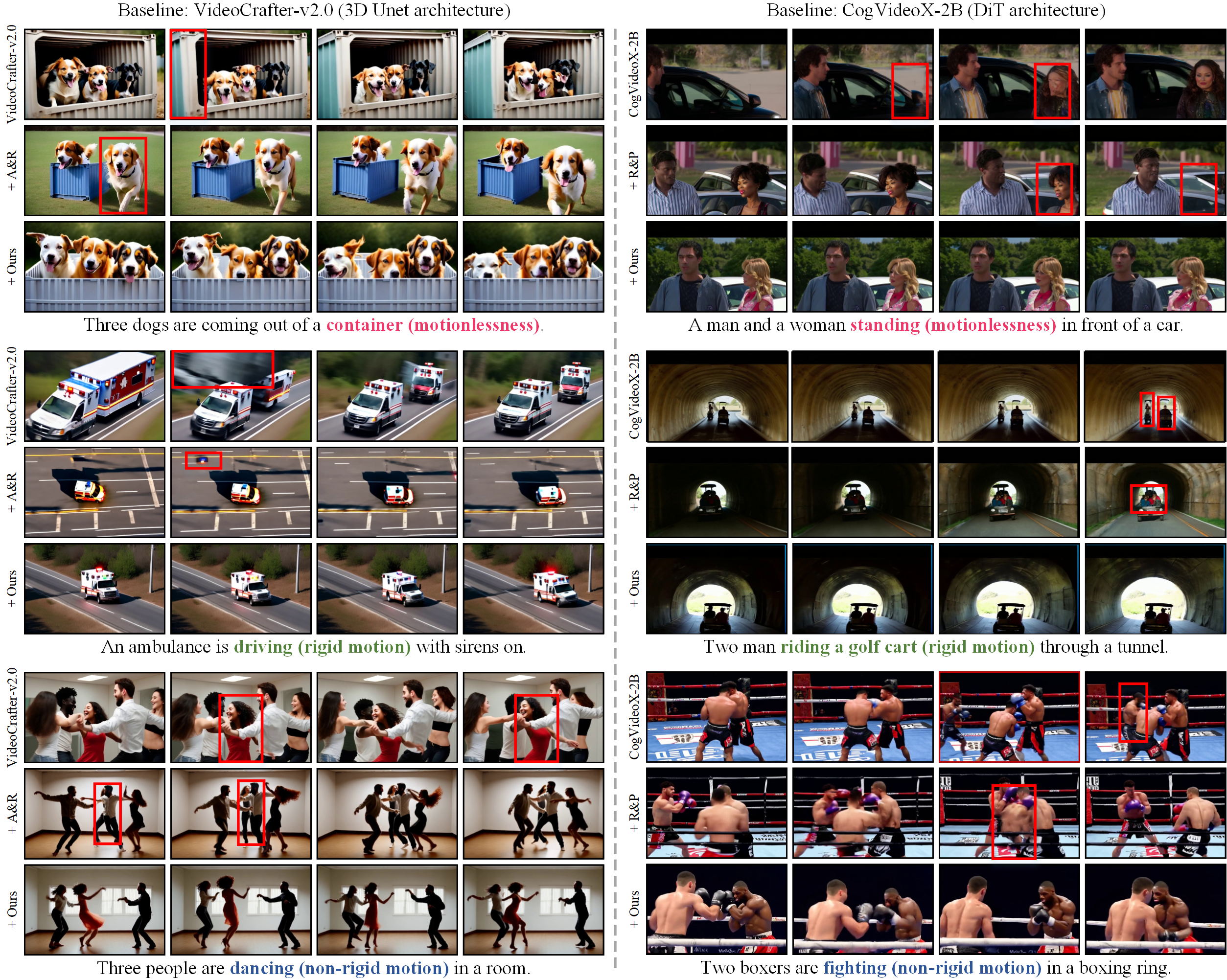}
    \caption{Visual cases: comparisons under diverse motion categories. 
    In 3D Unet architecture, we compare our framework with baseline VideoCrafter-v2.0 \cite{chen2024videocrafter2} and compositional approach A\&R \cite{phung2024grounded}. While in DiT architecture, we compare our framework with baseline CogVideoX-2B \cite{yang2024cogvideox} and compositional approach R\&P \cite{chen2024training}.
    Our framework yields improved cross-frame consistency and motion fidelity.}
    \label{fig:visualization}
\end{figure*}

\subsection{Experimental Setups}

\textbf{Benchmarks.} We construct new benchmarks for CVG performance evaluation, by selecting compositional video descriptions from real-world video datasets MSR-VTT \cite{xu2016msr} and Panda-70M \cite{chen2024panda}. To consider linguistic diversity, we categorize descriptions into four compositional modes: Coordinating Structure, Quantitative Expression, Collective Noun, and Interactive Verb, as detailed in {\color{cvprblue} Appendix B}. Guided by such categorization, we employ Llama-v3.3-70B \cite{Llama3.3} to identify samples with the above linguistic modes. As a result, we obtain two benchmarks: CVGBench-m (1665 samples from MSR-VTT) and CVGBench-p (994 samples from Panda-70M).

\noindent \textbf{Evaluation metrics.} We adopt five “\textit{Temporal Quality}” metrics, as defined in VBench \cite{huang2024vbench}: (1) Subject Consistency: measures consistency of a foreground instance's appearance by computing the similarity of DINO features \cite{caron2021emerging}; (2) Background Consistency: assesses coherence of the background scene by calculating the similarity of CLIP features \cite{radford2021learning}; (3) Temporal Flickering: is computed as the mean absolute difference between consecutive frames; (4) Motion Smoothness: evaluates continuity of generated motions based on motion priors from AMT model \cite{li2023amt}; and (5) Dynamic Degree: uses RAFT model \cite{teed2020raft} to assess whether a generated video has large motions.

\noindent \textbf{Baselines.} We compare our approach against several groups of open-source baseline models: (1) Traditional T2V models. These include LVDM \cite{he2022latent}, modelScope \cite{wang2023modelscope}, LATTE \cite{ma2024latte}, LaVie \cite{wang2024lavie}, Show-1 \cite{zhang2024show}, VideoCrafter-v1.0 \cite{chen2023videocrafter1},  VideoCrafter-v2.0  \cite{chen2024videocrafter2}, OpenSora-v1.2 \cite{zheng2024open}, T2V-Turbo-v2.0 \cite{li2024t2v}, and CogVideoX-2B \cite{yang2024cogvideox}; (2) Compositional T2V model. This includes VideoTetris \cite{tian2024videotetris} and Vico \cite{yang2024compositional}; (3) Compositional T2I model. We use VideoCrafter-v2.0 as the baseline to reproduce BoxDiff \cite{xie2023boxdiff}, R\&B \cite{xiao2023r}, and A\&R \cite{phung2024grounded}. Similarly, we use CogVideoX-2B as the baseline to reproduce R\&P \cite{chen2024training}.

\noindent \textbf{Implementation details.} For \textit{structured motion reasoning} module, we use LLaMA-v3.3-70B \cite{dubey2024llama} as a baseline model. For \textit{disentangled motion guidance} module, we apply this to both VideoCrafter-v2.0  \cite{chen2024videocrafter2} and CogVideoX-2B \cite{yang2024cogvideox}. VideoCrafter-v2.0 adopts Unet design, while CogVideoX-2B utilizes DiT architecture. Since motion guidance requires avoiding instance omission as a prerequisite, we incorporate A\&R \cite{phung2024grounded} and R\&P \cite{chen2024training} into VideoCrafter-v2.0 and into CogVideo-2B, respectively. The hyperparameters are set as follows: for VideoCrafter-v2.0, we set the guidance factor $\beta$ to 10 and apply motion guidance during denoising steps 1 to 25; for CogVideoX-2B, $\beta$ is set to 0.15 and the guidance is applied during steps 1 to 10.

\subsection{Quantitative Comparisons}
As shown in~\cref{tab:video_comparison}, we perform a comprehensive evaluation of various video generation models. In both VideoCrafter-v2.0 \cite{chen2024videocrafter2} (3D U-Net architecture) and CogVideoX-2B \cite{yang2024cogvideox} (DiT architecture) baselines, our framework achieves steady improvements across all five evaluation dimensions on the CVGBench-m and CVGBench-p benchmarks. For example, compared to the compositional visual generation approach R\&P \cite{chen2024training}, our framework can improve Subject Consistency from 91.00\% to 98.27\%, Background Consistency from 90.85\% to 97.73\%, Temporal Flickering from 95.07\% to 98.25\%, Motion Smoothness from 96.96\% to 98.74\%, and Dynamic Degree from 91.02\% to 96.00\%. However, R\&P sometimes compromises Subject and Background Consistency. This is because R\&P is designed to solve semantic leakage between instances in a frame-independent manner, but neglects to model the cross-frame consistency. Our framework is able to rectify motion categories in an instance independent manner. For static instances, pixel-wise consistency across frames is enforced; for moving ones, displacement follows predefined motion vectors.

\subsection{Qualitative Comparisons}
\noindent \textbf{Synthesizing basic motion categories.} As shown in~\cref{fig:visualization}, we provide visualization comparisons across diverse motion categories, including motionlessness, rigid motion, and non-rigid motion. Considering motionless scenarios (e.g., “static container”, or “man and woman standing”), our approach effectively suppresses undesired movement, enabled by its awareness of cross-frame feature consistency. Considering rigid motion examples (e.g., “ambulance driving”, or “riding a golf cart”), our method not only preserves global instance integrity but also enforces displacement. Other methods suffer from unnatural deformation or minimal motion. This is achieved by maintaining geometric invariance as instances move. Considering non-rigid motion cases (e.g., “people dancing”, or “boxer fighting”), our framework produces expressive body dynamics, while comparison methods fail to preserve coherent pose progression. This is because our framework can model complex deformation by pixel-wise motion fields.

\begin{figure}[tb!]
    \centering
    \includegraphics[width=\linewidth]{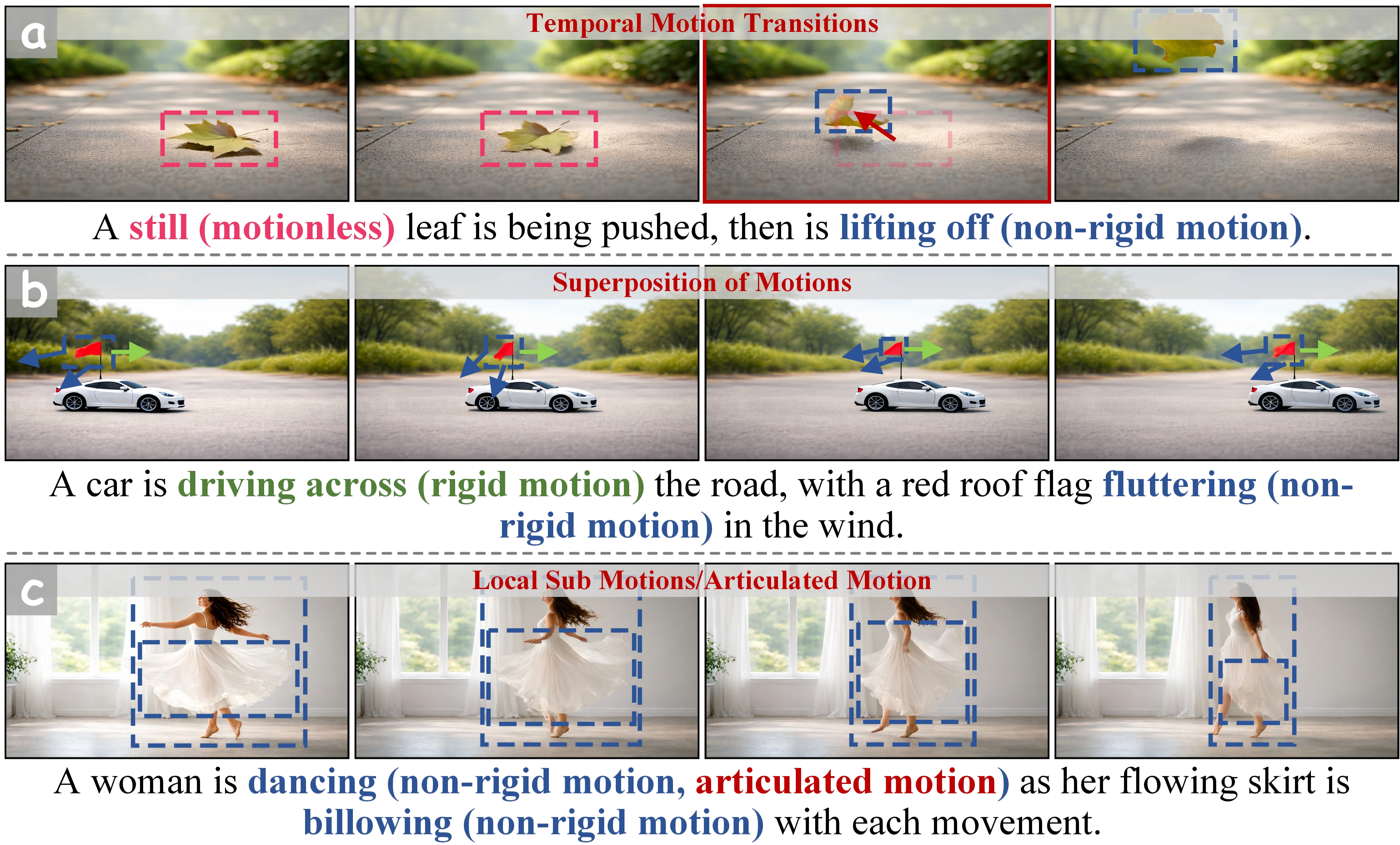} 
    \caption{Visual cases: generation of complex motion generation.}
    \label{diverse_motion}
\end{figure}
\noindent \textbf{Handling complex motion behaviors.} As shown in\cref{diverse_motion}, our framework can synthesize complex compositional motion behaviors formed by multiple categories of motion. In (a), a leaf changes from rest to non-rigid motion after being pushed, which enables by sequential application of diverse guidance modules across frames. In (b), generated video shows the coexistence of rigid-body car motion and non-rigid flag oscillation. This is achieved by spatial composition of diverse guidance modules with varying strengths. In (c), our framework can effectively capture local sub-motions in articulated human motion. This is realized by decomposing the instance into sub-regions, each with region-specific guidance.

\subsection{Ablation Study}
\begin{table}[tb!]
\centering
\caption{Ablation analysis of diverse backbones for motion reasoning. Best scores are \textbf{bolded}.}
\resizebox{\linewidth}{!}{%
\begin{tabular}{cccccc}
\toprule
 
\multirow{2}{*}{Backbones}& Subject & Background & Temporal & Motion & Dynamic  \\
 & Consistency & Consistency & Flickering & Smoothness & Degree \\
\midrule
\multicolumn{6}{c}{Baseline: VideoCrafter-v2.0 \cite{chen2024videocrafter2}} \\
LLaMA-8B \cite{dubey2024llama} & 97.55\% & 97.35\% & 96.55\% & 98.22\% & 75.34\% \\
LLaMA-70B \cite{Llama3.3} & \textbf{98.40\%} & \textbf{98.11\%} & \textbf{97.39\%} & \textbf{98.63\%} & \textbf{82.21\%} \\
\midrule
\multicolumn{6}{c}{Baseline: CogVideoX-2B \cite{yang2024cogvideox}} \\
LLaMA-8B \cite{dubey2024llama} & 96.70\% & 96.34\% & 97.32\% & 98.17\% & 94.77\% \\
LLaMA-70B \cite{Llama3.3} & \textbf{98.27\%} & \textbf{97.73\%} & \textbf{98.25\%} & \textbf{98.74\%} & \textbf{96.00\%} \\
\bottomrule
\end{tabular}
}
\label{tab:llama}
\end{table}

\noindent \textbf{Analysis of large language model scale.} To assess the impact of LLM scale on the generation of scene configuration, we compare the LLaMA-3.1-8B \cite{dubey2024llama} and LLaMA-3.3-70B \cite{Llama3.3} backbones. As shown in~\cref{tab:llama}, the 70B model significantly outperforms the 8B counterpart, especially on Subject Consistency (97.55\%$\rightarrow$98.40\%), Background Consistency (97.35\%$\rightarrow$98.11\%), and Dynamic Degree (75.34\%$\rightarrow$82.21\%). These highlight the superiority of stronger language model to reason frame-wise shapes and positions for each individual instance, ultimately improving video generation quality.

\newcommand{\cmark}{\ding{51}}
\newcommand{\xmark}{\ding{55}}

\begin{table}[t]
\centering
\caption{Ablation analysis of diverse motion guidance components, including Reference Conditioned Guidance (RCG), Geometric Invariance Guidance (GIG), Spatial Deformation Guidance (SDG). Best scores are \textbf{bolded}.}
\setlength{\heavyrulewidth}{1pt}
\resizebox{\linewidth}{!}{%
\begin{tabular}{cccccccc}
\toprule
\multirow{2}{*}{RCG} & \multirow{2}{*}{GIG} & \multirow{2}{*}{SDG} & Subject & Background & Temporal & Motion & Dynamic    \\
 &  & & Consistency & Consistency & Flickering & Smoothness & Degree  \\
\midrule
\multicolumn{8}{c}{Baseline: VideoCrafter-v2.0 \cite{chen2024videocrafter2}} \\
\xmark & \xmark & \xmark & 97.48\% & 97.05\% & 96.43\% & 98.27\% & 38.40\%   \\
\cmark & \xmark & \xmark & 98.11\% & 97.85\% & 97.21\% & 98.55\% & 51.60\% \\
\xmark & \cmark & \xmark & 98.07\% & 97.80\% & 97.15\% & 98.52\% & 53.60\% \\
\xmark & \xmark & \cmark & 97.71\% & 97.40\% & 96.72\% & 98.36\% & 74.85\% \\
\cmark & \cmark & \xmark & 98.15\% & 97.91\% & 97.23\% & 98.57\% & 54.40\% \\
\cmark & \cmark & \cmark & \textbf{98.40\%} & \textbf{98.11\%} & \textbf{97.39\%} & \textbf{98.63\%} & \textbf{82.21\%} \\
\midrule
\multicolumn{8}{c}{Baseline: CogVideoX-2B \cite{yang2024cogvideox}} \\
\xmark & \xmark & \xmark & 91.00\% & 90.85\% & 95.07\% & 96.96\% & 91.02\%  \\
\cmark & \xmark & \xmark & 95.98\% & 96.03\% & 96.89\% & 97.90\% & 92.16\% \\
\xmark & \cmark & \xmark & 96.21\% & 96.07\% & 96.88\% & 97.88\% & 92.77\% \\
\xmark & \xmark & \cmark & 96.13\% & 96.14\% & 96.83\% & 97.81\% & 93.57\% \\
\cmark & \cmark & \xmark & 97.19\% & 96.91\% & 97.49\% & 98.26\% & 94.56\% \\
\cmark & \cmark & \cmark & \textbf{98.27\%} & \textbf{97.73\%} & \textbf{98.25\%} & \textbf{98.74\%} & \textbf{96.00\%} \\
\bottomrule
\end{tabular}
}
\label{tab:guidance_ablation}
\end{table}

\noindent \textbf{Analysis of diverse guidance branches.} As shown in~\cref{tab:guidance_ablation}, progressively incorporating each guidance branch consistently improves all evaluation metrics. These gains arise from two factors. On one hand, all guidance enhances cross-frame feature propagation within foreground regions. For example, when using CogVideoX-2B \cite{yang2024cogvideox} as baseline, our framework achieves a $\sim$5.0\% improvement in terms of cross-frame consistency. On the other hand, rigid and non-rigid motion guidance branches enforce video generation model to synthesize large-scale motion specified by motion representations. For instance, with VideoCrafter-v2.0 \cite{chen2024videocrafter2} as baseline, our non-rigid motion guidance yields Dynamic Degree gain of 27.81\%.


\subsection{Failure Cases}
\begin{figure}[tb!]
    \centering
    \includegraphics[width=\linewidth]{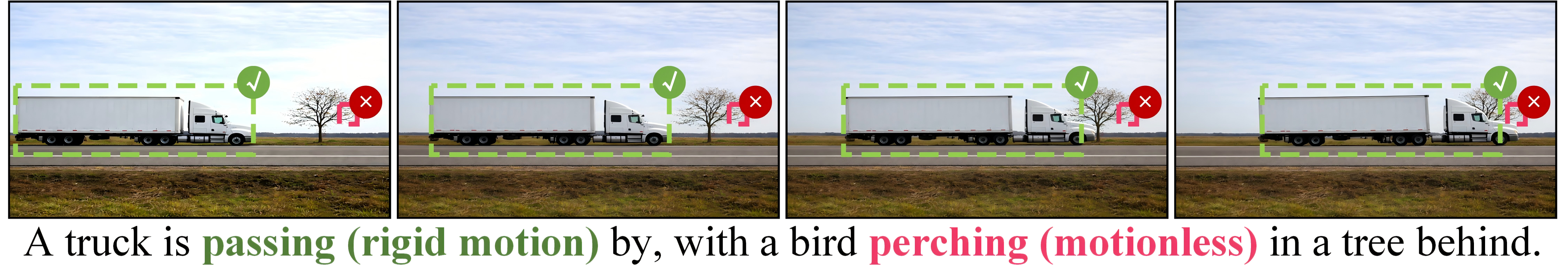} 
    \caption{Failure case: Overly small bounding boxes are insufficient for guiding object (motionlessness) generation.}
    \label{failure_case}
\end{figure}

While our framework can improve motion diversity in video generation, challenges remain in handling small objects with insufficient motion evidence. As shown in~\cref{failure_case}, a large vehicle is successfully generated, whereas the small bird is not faithfully reproduced in the generated video. This is because overly small boxes provide insufficient spatial coverage for effective guidance. One promising direction is to use scale-aware guidance \cite{ma2023trailblazer}.
\section{Conclusion}
This paper proposes a motion factorization framework, enhancing motion diversity in compositional video generation without additional learning. The key idea is to decompose complex scene dynamics into diverse categories, including motionlessness, rigid motion, non-rigid motion. Thus, each motion category can be independently modeled and guided. Specifically, we \textit{resolve semantic ambiguities} by reformulating user-provided prompts as a motion graph of instances and their interactions. This enables reliable reasoning over motion representations of individual instances. Then, we \textit{address motion homogenization} by separately stabilizing background appearance, preserving rigid-body geometry, and regularizing non-rigid deformation. Experiments demonstrate the effectiveness of our framework in generating desired motion behaviors. However, our framework assumes stable viewpoints and simply focuses on instance-wise motion factorization. Our future work will explore camera poses to model global viewpoint changes across scenes \cite{he2025cameractrl,bahmani2025ac3d}.

\noindent \textbf{Acknowledgement:} This work was supported by the National Natural Science Foundation of China (No. U23B2013, 62276176). This work was also partly supported by the SICHUAN Provincial Natural Science Foundation (No. 2024NSFJQ0023).

\balance
{
    \small
    \bibliographystyle{ieeenat_fullname}
    \bibliography{main}

@String(IJCV = {Int. J. Comput. Vis.})

@String(CVPR= {IEEE Conf. Comput. Vis. Pattern Recog.})

@String(ICCV= {Int. Conf. Comput. Vis.})

@String(ECCV= {Eur. Conf. Comput. Vis.})

@String(TIP  = {IEEE Trans. Image Process.})

@String(ICLR = {Int. Conf. Learn. Represent.})

@String(IJCV  = {IJCV})

@String(CVPR  = {CVPR})

@String(ICCV  = {ICCV})

@String(ECCV  = {ECCV})

@String(TIP   = {IEEE TIP})

@String(ICLR  = {ICLR})

@article{wang2024lavie,
  title={Lavie: High-quality video generation with cascaded latent diffusion models},
  author={Wang, Yaohui and Chen, Xinyuan and Ma, Xin and Zhou, Shangchen and Huang, Ziqi and Wang, Yi and Yang, Ceyuan and He, Yinan and Yu, Jiashuo and Yang, Peiqing and others},
  journal={International Journal of Computer Vision (IJCV)},
  pages={1--20},
  year={2024},
  publisher={Springer}
}

@inproceedings{xu2016msr,
  title={MSR-VTT: A large video description dataset for bridging video and language},
  author={Xu, Jun and Mei, Tao and Yao, Ting and Rui, Yong},
  booktitle={Proceedings of the IEEE Conference on Computer Vision and Pattern Recognition (CVPR)},
  pages={5288--5296},
  year={2016}
}

@inproceedings{chen2024pixart,
  title={Pixart-$\sigma$: Weak-to-strong training of diffusion transformer for 4k text-to-image generation},
  author={Chen, Junsong and Ge, Chongjian and Xie, Enze and Wu, Yue and Yao, Lewei and Ren, Xiaozhe and Wang, Zhongdao and Luo, Ping and Lu, Huchuan and Li, Zhenguo},
  booktitle={European Conference on Computer Vision (ECCV)},
  pages={74--91},
  year={2024},
  organization={Springer}
}

@article{li2024hunyuan,
  title={Hunyuan-dit: A powerful multi-resolution diffusion transformer with fine-grained chinese understanding},
  author={Li, Zhimin and Zhang, Jianwei and Lin, Qin and Xiong, Jiangfeng and Long, Yanxin and Deng, Xinchi and Zhang, Yingfang and Liu, Xingchao and Huang, Minbin and Xiao, Zedong and others},
  journal={arXiv preprint arXiv:2405.08748},
  year={2024}
}

@inproceedings{esser2024scaling,
  title={Scaling rectified flow transformers for high-resolution image synthesis},
  author={Esser, Patrick and Kulal, Sumith and Blattmann, Andreas and Entezari, Rahim and M{\"u}ller, Jonas and Saini, Harry and Levi, Yam and Lorenz, Dominik and Sauer, Axel and Boesel, Frederic and others},
  booktitle={International Conference on Machine Learning (ICML)},
  year={2024}
}

@inproceedings{xie2023boxdiff,
  title={Boxdiff: Text-to-image synthesis with training-free box-constrained diffusion},
  author={Xie, Jinheng and Li, Yuexiang and Huang, Yawen and Liu, Haozhe and Zhang, Wentian and Zheng, Yefeng and Shou, Mike Zheng},
  booktitle={Proceedings of the IEEE/CVF International Conference on Computer Vision (ICCV)},
  pages={7452--7461},
  year={2023}
}

@article{xiao2023r,
  title={R\&b: Region and boundary aware zero-shot grounded text-to-image generation},
  author={Xiao, Jiayu and Lv, Henglei and Li, Liang and Wang, Shuhui and Huang, Qingming},
  journal={arXiv preprint arXiv:2310.08872},
  year={2023}
}

@inproceedings{phung2024grounded,
  title={Grounded text-to-image synthesis with attention refocusing},
  author={Phung, Quynh and Ge, Songwei and Huang, Jia-Bin},
  booktitle={Proceedings of the IEEE/CVF Conference on Computer Vision and Pattern Recognition (CVPR)},
  pages={7932--7942},
  year={2024}
}

@article{zhang2025magiccomp,
  title={MagicComp: Training-free Dual-Phase Refinement for Compositional Video Generation},
  author={Zhang, Hongyu and Deng, Yufan and Yuan, Shenghai and Jin, Peng and Cheng, Zesen and Zhao, Yian and Liu, Chang and Chen, Jie},
  journal={arXiv preprint arXiv:2503.14428},
  year={2025}
}

@inproceedings{jeong2024vmc,
  title={Vmc: Video motion customization using temporal attention adaption for text-to-video diffusion models},
  author={Jeong, Hyeonho and Park, Geon Yeong and Ye, Jong Chul},
  booktitle={Proceedings of the IEEE/CVF Conference on Computer Vision and Pattern Recognition (CVPR)},
  pages={9212--9221},
  year={2024}
}

@inproceedings{chen2024panda,
  title={Panda-70m: Captioning 70m videos with multiple cross-modality teachers},
  author={Chen, Tsai-Shien and Siarohin, Aliaksandr and Menapace, Willi and Deyneka, Ekaterina and Chao, Hsiang-wei and Jeon, Byung Eun and Fang, Yuwei and Lee, Hsin-Ying and Ren, Jian and Yang, Ming-Hsuan and others},
  booktitle={Proceedings of the IEEE/CVF Conference on Computer Vision and Pattern Recognition (CVPR)},
  pages={13320--13331},
  year={2024}
}

@inproceedings{chen2024videocrafter2,
  title={Videocrafter2: Overcoming data limitations for high-quality video diffusion models},
  author={Chen, Haoxin and Zhang, Yong and Cun, Xiaodong and Xia, Menghan and Wang, Xintao and Weng, Chao and Shan, Ying},
  booktitle={Proceedings of the IEEE/CVF Conference on Computer Vision and Pattern Recognition (CVPR)},
  pages={7310--7320},
  year={2024}
}

@article{yang2024cogvideox,
  title={Cogvideox: Text-to-video diffusion models with an expert transformer},
  author={Yang, Zhuoyi and Teng, Jiayan and Zheng, Wendi and Ding, Ming and Huang, Shiyu and Xu, Jiazheng and Yang, Yuanming and Hong, Wenyi and Zhang, Xiaohan and Feng, Guanyu and others},
  journal={arXiv preprint arXiv:2408.06072},
  year={2024}
}

@article{he2022latent,
  title={Latent video diffusion models for high-fidelity long video generation},
  author={He, Yingqing and Yang, Tianyu and Zhang, Yong and Shan, Ying and Chen, Qifeng},
  journal={arXiv preprint arXiv:2211.13221},
  year={2022}
}

@inproceedings{wu2024lamp,
  title={Lamp: Learn a motion pattern for few-shot video generation},
  author={Wu, Ruiqi and Chen, Liangyu and Yang, Tong and Guo, Chunle and Li, Chongyi and Zhang, Xiangyu},
  booktitle={Proceedings of the IEEE/CVF Conference on Computer Vision and Pattern Recognition (CVPR)},
  pages={7089--7098},
  year={2024}
}

@article{ma2024latte,
  title={Latte: Latent diffusion transformer for video generation},
  author={Ma, Xin and Wang, Yaohui and Jia, Gengyun and Chen, Xinyuan and Liu, Ziwei and Li, Yuan-Fang and Chen, Cunjian and Qiao, Yu},
  journal={arXiv preprint arXiv:2401.03048},
  year={2024}
}

@article{zheng2024open,
  title={Open-sora: Democratizing efficient video production for all},
  author={Zheng, Zangwei and Peng, Xiangyu and Yang, Tianji and Shen, Chenhui and Li, Shenggui and Liu, Hongxin and Zhou, Yukun and Li, Tianyi and You, Yang},
  journal={arXiv preprint arXiv:2412.20404},
  year={2024}
}

@inproceedings{tian2024videotetris,
  title={Videotetris: Towards compositional text-to-video generation},
  author={Tian, Ye and Yang, Ling and Yang, Haotian and Gao, Yuan and Deng, Yufan and Wang, Xintao and Yu, Zhaochen and Tao, Xin and Wan, Pengfei and ZHANG, Di and others},
  booktitle={Advances in Neural Information Processing Systems (NeurIPS)},
  volume={37},
  pages={29489--29513},
  year={2024}
}

@article{yang2024compositional,
  title={Compositional video generation as flow equalization},
  author={Yang, Xingyi and Wang, Xinchao},
  journal={arXiv preprint arXiv:2407.06182},
  year={2024}
}

@inproceedings{lian2023llm,
  title={Llm-grounded video diffusion models},
  author={Lian, Long and Shi, Baifeng and Yala, Adam and Darrell, Trevor and Li, Boyi},
  booktitle={International Conference on Learning Representations (ICLR)},
  year={2024}
}

@article{lin2023videodirectorgpt,
  title={Videodirectorgpt: Consistent multi-scene video generation via llm-guided planning},
  author={Lin, Han and Zala, Abhay and Cho, Jaemin and Bansal, Mohit},
  journal={Conference on Language Modeling (CoLM)},
  year={2024}
}

@article{geyer2023tokenflow,
  title={Tokenflow: Consistent diffusion features for consistent video editing},
  author={Geyer, Michal and Bar-Tal, Omer and Bagon, Shai and Dekel, Tali},
  journal={arXiv preprint arXiv:2307.10373},
  year={2023}
}

@article{ma2023trailblazer,
  title={TrailBlazer: Trajectory Control for Diffusion-Based Video Generation},
  author={Ma, Wan-Duo Kurt and Lewis, JP and Kleijn, W Bastiaan},
  journal={arXiv preprint arXiv:2401.00896},
  year={2023}
}

@article{qiu2024freetraj,
  title={Freetraj: Tuning-free trajectory control in video diffusion models},
  author={Qiu, Haonan and Chen, Zhaoxi and Wang, Zhouxia and He, Yingqing and Xia, Menghan and Liu, Ziwei},
  journal={arXiv preprint arXiv:2406.16863},
  year={2024}
}

@inproceedings{koroglu2025onlyflow,
  title={Onlyflow: Optical flow based motion conditioning for video diffusion models},
  author={Koroglu, Mathis and Caselles-Dupr{\'e}, Hugo and Jeanneret, Guillaume and Cord, Matthieu},
  booktitle={Proceedings of the IEEE/CVF Conference on Computer Vision and Pattern Recognition (CVPR)},
  pages={6226--6236},
  year={2025}
}

@inproceedings{zhang2025tora,
  title={Tora: Trajectory-oriented diffusion transformer for video generation},
  author={Zhang, Zhenghao and Liao, Junchao and Li, Menghao and Dai, Zuozhuo and Qiu, Bingxue and Zhu, Siyu and Qin, Long and Wang, Weizhi},
  booktitle={Proceedings of the IEEE/CVF Conference on Computer Vision and Pattern Recognition (CVPR)},
  pages={2063--2073},
  year={2025}
}

@article{yin2023dragnuwa,
  title={Dragnuwa: Fine-grained control in video generation by integrating text, image, and trajectory},
  author={Yin, Shengming and Wu, Chenfei and Liang, Jian and Shi, Jie and Li, Houqiang and Ming, Gong and Duan, Nan},
  journal={arXiv preprint arXiv:2308.08089},
  year={2023}
}

@inproceedings{burgert2025go,
  title={Go-with-the-flow: Motion-controllable video diffusion models using real-time warped noise},
  author={Burgert, Ryan and Xu, Yuancheng and Xian, Wenqi and Pilarski, Oliver and Clausen, Pascal and He, Mingming and Ma, Li and Deng, Yitong and Li, Lingxiao and Mousavi, Mohsen and others},
  booktitle={Proceedings of the IEEE/CVF Conference on Computer Vision and Pattern Recognition (CVPR)},
  pages={13--23},
  year={2025}
}

@inproceedings{geng2025motion,
  title={Motion prompting: Controlling video generation with motion trajectories},
  author={Geng, Daniel and Herrmann, Charles and Hur, Junhwa and Cole, Forrester and Zhang, Serena and Pfaff, Tobias and Lopez-Guevara, Tatiana and Aytar, Yusuf and Rubinstein, Michael and Sun, Chen and others},
  booktitle={Proceedings of the IEEE/CVF Conference on Computer Vision and Pattern Recognition (CVPR)},
  pages={1--12},
  year={2025}
}

@article{wang2023modelscope,
  title={Modelscope text-to-video technical report},
  author={Wang, Jiuniu and Yuan, Hangjie and Chen, Dayou and Zhang, Yingya and Wang, Xiang and Zhang, Shiwei},
  journal={arXiv preprint arXiv:2308.06571},
  year={2023}
}

@article{zhang2024show,
  title={Show-1: Marrying pixel and latent diffusion models for text-to-video generation},
  author={Zhang, David Junhao and Wu, Jay Zhangjie and Liu, Jia-Wei and Zhao, Rui and Ran, Lingmin and Gu, Yuchao and Gao, Difei and Shou, Mike Zheng},
  journal={International Journal of Computer Vision (IJCV)},
  pages={1--15},
  year={2024},
  publisher={Springer}
}

@article{li2024t2v,
  title={T2v-turbo-v2: Enhancing video generation model post-training through data, reward, and conditional guidance design},
  author={Li, Jiachen and Long, Qian and Zheng, Jian and Gao, Xiaofeng and Piramuthu, Robinson and Chen, Wenhu and Wang, William Yang},
  journal={arXiv preprint arXiv:2410.05677},
  year={2024}
}

@misc{pika2024,
  author       = {Pika},
  title        = {Pika},
  howpublished = {\url{https://www.pika.art}},
  year         = {2024},
}

@misc{Runway2024,
  author       = {Runway},
  title        = {Introducing gen-3 alpha: A new frontier for video generation},
  howpublished = {\url{https://runwayml.com/research/introducing-gen-3-alpha}},
  year         = {2024},
}

@misc{dreamina2024,
  author       = {Capcut},
  title        = {Dreamina},
  howpublished = {\url{https://dreamina.capcut.com/ai-tool/home}},
  year         = {2024},
}

@misc{kling2024,
  author       = {Kuaishou},
  title        = {Kling},
  howpublished = {\url{https://kling.kuaishou.com/}},
  year         = {2024},
}

@article{chen2023videocrafter1,
  title={Videocrafter1: Open diffusion models for high-quality video generation},
  author={Chen, Haoxin and Xia, Menghan and He, Yingqing and Zhang, Yong and Cun, Xiaodong and Yang, Shaoshu and Xing, Jinbo and Liu, Yaofang and Chen, Qifeng and Wang, Xintao and others},
  journal={arXiv preprint arXiv:2310.19512},
  year={2023}
}

@article{chen2024training,
  title={Training-free regional prompting for diffusion transformers},
  author={Chen, Anthony and Xu, Jianjin and Zheng, Wenzhao and Dai, Gaole and Wang, Yida and Zhang, Renrui and Wang, Haofan and Zhang, Shanghang},
  journal={arXiv preprint arXiv:2411.02395},
  year={2024}
}

@inproceedings{caron2021emerging,
  title={Emerging properties in self-supervised vision transformers},
  author={Caron, Mathilde and Touvron, Hugo and Misra, Ishan and J{\'e}gou, Herv{\'e} and Mairal, Julien and Bojanowski, Piotr and Joulin, Armand},
  booktitle={Proceedings of the IEEE/CVF International Conference on Computer Vision (ICCV)},
  pages={9650--9660},
  year={2021}
}

@inproceedings{radford2021learning,
  title={Learning transferable visual models from natural language supervision},
  author={Radford, Alec and Kim, Jong Wook and Hallacy, Chris and Ramesh, Aditya and Goh, Gabriel and Agarwal, Sandhini and Sastry, Girish and Askell, Amanda and Mishkin, Pamela and Clark, Jack and others},
  booktitle={International Conference on Machine Learning (ICML)},
  pages={8748--8763},
  year={2021},
  organization={PmLR}
}

@inproceedings{li2023amt,
  title={Amt: All-pairs multi-field transforms for efficient frame interpolation},
  author={Li, Zhen and Zhu, Zuo-Liang and Han, Ling-Hao and Hou, Qibin and Guo, Chun-Le and Cheng, Ming-Ming},
  booktitle={Proceedings of the IEEE/CVF Conference on Computer Vision and Pattern Recognition (CVPR)},
  pages={9801--9810},
  year={2023}
}

@inproceedings{teed2020raft,
  title={Raft: Recurrent all-pairs field transforms for optical flow},
  author={Teed, Zachary and Deng, Jia},
  booktitle={European Conference on Computer Vision (ECCV)},
  pages={402--419},
  year={2020},
  organization={Springer}
}

@inproceedings{huang2024vbench,
  title={Vbench: Comprehensive benchmark suite for video generative models},
  author={Huang, Ziqi and He, Yinan and Yu, Jiashuo and Zhang, Fan and Si, Chenyang and Jiang, Yuming and Zhang, Yuanhan and Wu, Tianxing and Jin, Qingyang and Chanpaisit, Nattapol and others},
  booktitle={Proceedings of the IEEE/CVF Conference on Computer Vision and Pattern Recognition (CVPR)},
  pages={21807--21818},
  year={2024}
}

@article{ZeroScope2023,
  title={ZeroScope},
  year={2023}
}

@article{Llama3.3,
  title={Llama 3.3},
  year={2024},
  url={https://www.llama.com/docs/model-cards-and-prompt-formats/llama3_3/}
}

@article{jin2021subjective,
  title={Subjective and objective quality assessment of 2d and 3d foveated video compression in virtual reality},
  author={Jin, Yize and Chen, Meixu and Goodall, Todd and Patney, Anjul and Bovik, Alan C},
  journal={IEEE Transactions on Image Processing (IEEE TIP)},
  volume={30},
  pages={5905--5919},
  year={2021},
  publisher={IEEE}
}

@article{zuo2023fine,
  title={Fine-grained video retrieval with scene sketches},
  author={Zuo, Ran and Deng, Xiaoming and Chen, Keqi and Zhang, Zhengming and Lai, Yu-Kun and Liu, Fang and Ma, Cuixia and Wang, Hao and Liu, Yong-Jin and Wang, Hongan},
  journal={IEEE Transactions on Image Processing (IEEE TIP)},
  volume={32},
  pages={3136--3149},
  year={2023},
  publisher={IEEE}
}

@article{dubey2024llama,
  title={The llama 3 herd of models},
  author={Dubey, Abhimanyu and Jauhri, Abhinav and Pandey, Abhinav and Kadian, Abhishek and Al-Dahle, Ahmad and Letman, Aiesha and Mathur, Akhil and Schelten, Alan and Yang, Amy and Fan, Angela and others},
  journal={arXiv e-prints},
  pages={arXiv--2407},
  year={2024}
}

@article{huang2024genmac,
  title={Genmac: compositional text-to-video generation with multi-agent collaboration},
  author={Huang, Kaiyi and Huang, Yukun and Ning, Xuefei and Lin, Zinan and Wang, Yu and Liu, Xihui},
  journal={arXiv preprint arXiv:2412.04440},
  year={2024}
}

@article{li2024survey,
  title={A survey on long video generation: Challenges, methods, and prospects},
  author={Li, Chengxuan and Huang, Di and Lu, Zeyu and Xiao, Yang and Pei, Qingqi and Bai, Lei},
  journal={arXiv preprint arXiv:2403.16407},
  year={2024}
}

@article{lei2024comprehensive,
  title={A comprehensive survey on human video generation: Challenges, methods, and insights},
  author={Lei, Wentao and Wang, Jinting and Ma, Fengji and Huang, Guanjie and Liu, Li},
  journal={arXiv preprint arXiv:2407.08428},
  year={2024}
}

@article{xue2025human,
  title={Human motion video generation: A survey},
  author={Xue, Haiwei and Luo, Xiangyang and Hu, Zhanghao and Zhang, Xin and Xiang, Xunzhi and Dai, Yuqin and Liu, Jianzhuang and Zhang, Zhensong and Li, Minglei and Yang, Jian and others},
  journal={IEEE Transactions on Pattern Analysis and Machine Intelligence (TPAMI)},
  year={2025},
  publisher={IEEE}
}

@article{zhao2023survey,
  title={A survey of large language models},
  author={Zhao, Wayne Xin and Zhou, Kun and Li, Junyi and Tang, Tianyi and Wang, Xiaolei and Hou, Yupeng and Min, Yingqian and Zhang, Beichen and Zhang, Junjie and Dong, Zican and others},
  journal={arXiv preprint arXiv:2303.18223},
  year={2023}
}

@article{minaee2024large,
  title={Large language models: A survey},
  author={Minaee, Shervin and Mikolov, Tomas and Nikzad, Narjes and Chenaghlu, Meysam and Socher, Richard and Amatriain, Xavier and Gao, Jianfeng},
  journal={arXiv preprint arXiv:2402.06196},
  year={2024}
}

@article{hong2023direct2v,
  title={Direct2v: Large language models are frame-level directors for zero-shot text-to-video generation},
  author={Hong, Susung and Seo, Junyoung and Shin, Heeseong and Hong, Sunghwan and Kim, Seungryong},
  journal={arXiv preprint arXiv:2305.14330},
  year={2023}
}

@article{huang2023free,
  title={Free-bloom: Zero-shot text-to-video generator with llm director and ldm animator},
  author={Huang, Hanzhuo and Feng, Yufan and Shi, Cheng and Xu, Lan and Yu, Jingyi and Yang, Sibei},
  journal={Advances in Neural Information Processing Systems (NeurIPS)},
  volume={36},
  pages={26135--26158},
  year={2023}
}

@inproceedings{oh2024mevg,
  title={Mevg: Multi-event video generation with text-to-video models},
  author={Oh, Gyeongrok and Jeong, Jaehwan and Kim, Sieun and Byeon, Wonmin and Kim, Jinkyu and Kim, Sungwoong and Kim, Sangpil},
  booktitle={European Conference on Computer Vision (ECCV)},
  pages={401--418},
  year={2024},
  organization={Springer}
}

@article{he2025dyst,
  title={DyST-XL: Dynamic Layout Planning and Content Control for Compositional Text-to-Video Generation},
  author={He, Weijie and Liu, Mushui and Yu, Yunlong and Wang, Zhao and Wu, Chao},
  journal={arXiv preprint arXiv:2504.15032},
  year={2025}
}

@article{lu2023flowzero,
  title={Flowzero: Zero-shot text-to-video synthesis with llm-driven dynamic scene syntax},
  author={Lu, Yu and Zhu, Linchao and Fan, Hehe and Yang, Yi},
  journal={arXiv preprint arXiv:2311.15813},
  year={2023}
}

@inproceedings{zhuang2024vlogger,
  title={Vlogger: Make your dream a vlog},
  author={Zhuang, Shaobin and Li, Kunchang and Chen, Xinyuan and Wang, Yaohui and Liu, Ziwei and Qiao, Yu and Wang, Yali},
  booktitle={Proceedings of the IEEE/CVF Conference on Computer Vision and Pattern Recognition (CVPR)},
  pages={8806--8817},
  year={2024}
}

@inproceedings{liang2024movideo,
  title={Movideo: Motion-aware video generation with diffusion model},
  author={Liang, Jingyun and Fan, Yuchen and Zhang, Kai and Timofte, Radu and Van Gool, Luc and Ranjan, Rakesh},
  booktitle={European Conference on Computer Vision (ECCV)},
  pages={56--74},
  year={2024},
  organization={Springer}
}

@inproceedings{he2025cameractrl,
  title={Cameractrl II: Dynamic scene exploration via camera-controlled video diffusion models},
  author={He, Hao and Yang, Ceyuan and Lin, Shanchuan and Xu, Yinghao and Wei, Meng and Gui, Liangke and Zhao, Qi and Wetzstein, Gordon and Jiang, Lu and Li, Hongsheng},
  booktitle={Proceedings of the IEEE/CVF Conference on Computer Vision and Pattern Recognition (CVPR)},
  pages={13416--13426},
  year={2025}
}

@inproceedings{bahmani2025ac3d,
  title={Ac3d: Analyzing and improving 3d camera control in video diffusion transformers},
  author={Bahmani, Sherwin and Skorokhodov, Ivan and Qian, Guocheng and Siarohin, Aliaksandr and Menapace, Willi and Tagliasacchi, Andrea and Lindell, David B and Tulyakov, Sergey},
  booktitle={Proceedings of the IEEE/CVF Conference on Computer Vision and Pattern Recognition (CVPR)},
  pages={22875--22889},
  year={2025}
}
}

\end{document}